\documentclass[conference]{IEEEtran}
\usepackage{times}
\usepackage{graphicx}

\usepackage[numbers]{natbib}
\usepackage{multicol}
\usepackage[bookmarks=true]{hyperref}
\usepackage{amssymb}
\usepackage{amsmath}
\usepackage{booktabs}
\usepackage{multirow}
\usepackage{xcolor}

\newcommand{\vla}{ProgressVLA}

\newcommand{\Tau}{\mathrm{T}}

\begin{document}

\title{Progress-Guided Diffusion Policy for Vision-Language Robotic Manipulation}

\author{
\IEEEauthorblockN{Hongyu Yan$^{1}$, Qiwei Li$^{1}$, Jiaolong Yang$^{2}$, and Yadong Mu$^{1}$}
\IEEEauthorblockA{$^{1}$Peking University \hspace{1.5em} $^{2}$Microsoft Research Asia (MSRA)}
}



%

\twocolumn[{
\begin{@twocolumnfalse}
\maketitle
\vspace{-0.25in}
\begin{center}
\refstepcounter{figure}\label{fig:teaser}
\IfFileExists{figure1.pdf}{%
  \includegraphics[width=0.97\linewidth]{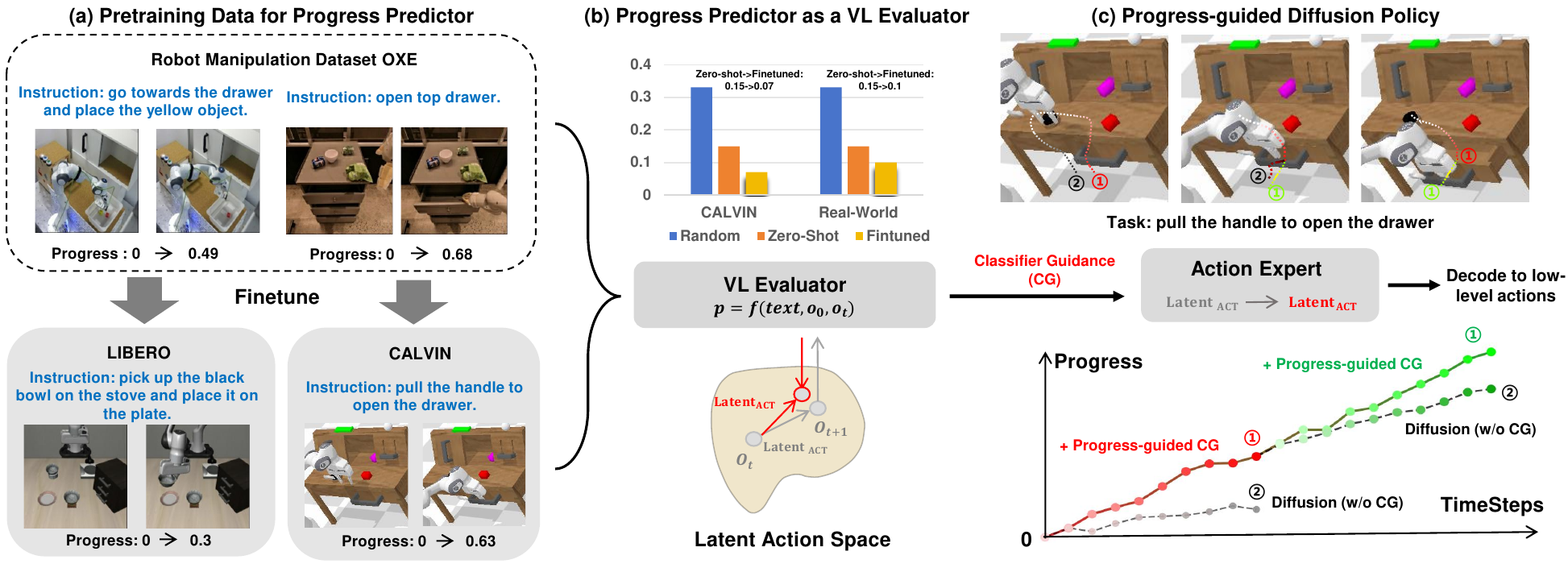}%
}{%
  \IfFileExists{Figure1.png}{%
    \includegraphics[width=0.98\linewidth]{Figure1.png}%
  }{%
    \setlength{\fboxsep}{6pt}%
    \fbox{%
      \begin{minipage}[c][1.6in][c]{0.98\textwidth}
        \centering
        \textbf{\figurename~\thefigure\ not found.}
      \end{minipage}%
    }%
  }%
}
\newline
{\footnotesize \figurename~\thefigure. Illustration of the key ideas in \vla. (a) We pretrain a vision/language-conditioned progress estimator on Open X-Embodiment (OXE)~\citep{o2024open} and finetune it on LIBERO~\citep{liu2023libero} / CALVIN~\citep{mees2022calvin}. (b) The estimator serves as a vision-language evaluator and achieves low residual after being finetuned (0.07 on CALVIN and 0.1 on real-world scenarios with a progress scale of $[0,1]$). (c) At inference, we use classifier (estimator) guidance in latent action space to steer diffusion toward higher progress. The refined latents are then decoded into action chunks for execution, often producing faster progress.}
\vspace{0.10in}
\end{center}

\end{@twocolumnfalse}
}]

\IEEEpeerreviewmaketitle

\section{Abstract}
\label{sec:abstract}


Most existing vision-language-action (VLA) models for robotic manipulation lack progress awareness, typically relying on hand-crafted heuristics for task termination. This limitation is particularly severe in long-horizon tasks involving cascaded sub-goals. In this work, we investigate the estimation and integration of task progress, proposing a novel model named {\textbf \vla}. Our technical contributions are twofold: (1) \emph{robust progress estimation}: We pre-train a progress estimator on large-scale, unsupervised video-text robotic datasets. This estimator achieves a low prediction residual (0.07 on a scale of $[0, 1]$) in simulation and demonstrates zero-shot generalization to unseen real-world samples, and (2) \emph{differentiable progress guidance}: We introduce an inverse dynamics world model that maps predicted action tokens into future latent visual states. These latents are then processed by the progress estimator; by applying a maximal progress regularization, we establish a differentiable pipeline that provides progress-piloted guidance to refine action tokens. Extensive experiments on the CALVIN and LIBERO benchmarks, alongside real-world robot deployment, consistently demonstrate substantial improvements in success rates and generalization over strong baselines.


\section{Introduction}

Recent vision-language-action (VLA) models have advanced policy learning by scaling to large robotics datasets~\citep{o2024open,walke2023bridgedata,khazatsky2024droid,bu2025agibot}. Yet many approaches still rely on dense action supervision~\cite{kim2024openvla}, which limits their scalability, while others depend on implicit and often noisy goal-satisfaction cues. Generative planners including tokenized autoregressive models~\citep{pertsch2025fast,bu2025univla} and diffusion policies~\cite{black2024pi_0,intelligence2504pi0}, can produce plausible trajectories. However, their sampling is largely driven by conditioning and typically lacks an explicit, dense notion of task progress. Consequently, long-horizon execution often relies on brittle termination heuristics rather than goal-directed generation. As a motivating fact, we present some empirical validation by us in Table~\ref{tab:teaser_intro}. It shows that explicitly guiding sampling with a learned progress signal substantially improves progress alignment and reduces the required steps for completing a task on CALVIN~\cite{mees2022calvin}, with a consistent gain in success.

\begin{table}[t]
  \centering
  \caption{\textbf{Key motivating observation.} Progress-guided sampling improves progress alignment and reduces completion steps on CALVIN.
  Pearson $r$ is the correlation between predicted progress $\{\hat p_t\}$ and a linear ramp $\{t/T\}$ over a rollout. Avg.\ steps denotes steps-to-completion, and Success is the task success rate. The baseline here is a standard diffusion policy for robotic manipulation.}
  \label{tab:teaser_intro}
  \setlength{\tabcolsep}{8pt}
  \begin{tabular}{cccc}
  \toprule
  Progress Guidance & \shortstack{Pearson $r\uparrow$} & Avg. steps$\downarrow$ & Success$\uparrow$ \\
  \midrule
   & 0.722 & 90.4 & 92.7 \\
  \checkmark & \textbf{0.934} & \textbf{77.3} & \textbf{93.6} \\
  \bottomrule
  \end{tabular}
  
\end{table}

To address these limitations, we propose a progress estimation technique (see Fig.~\ref{fig:teaser}) and integrate it to guide a diffusion policy, a framework we call \vla. Central to this approach is a progress estimator that outputs a normalized completion score, conditioned on the language-specified task and current visual observations. Why is progress estimation fundamental to long-horizon robotic manipulation? In vision-language conditioned tasks, a policy must transcend the generation of locally plausible motions; it must continuously evaluate whether its actions effectively advance toward the specified goal~\cite{intelligence2025pi}. Without a dense notion of progress, generative policies often squander computation on trajectories that appear visually reasonable but remain task-irrelevant. Furthermore, they lack a principled mechanism for termination, often defaulting to brittle, hand-crafted heuristics. However, learning progress directly from raw pixels is difficult~\cite{bu2025laof}. Robotic videos exhibit significant nuisance variations, such as camera jitter, background shifts, and distractor objects, which naive learning objectives often entangle with task dynamics. This results in progress signals that are noisy and poorly aligned with actual goal completion. \vla~mitigates this by grounding progress estimation within a pre-trained, object-centric visual feature space. By explicitly conditioning on language, the model ensures the learned signal prioritizes task-relevant state changes over incidental visual dynamics.

It is equally important to use progress during control. Rather than treating progress as a post-hoc evaluator, such as for reranking sampled trajectories or as a sparse success classifier~\cite{yu2025rlinf, zang2025rlinf, chen2025tgrpo, fei2025srpo}, \vla~embeds progress awareness directly into the action generation process. Specifically, for a given candidate action chunk, we develop an inverse dynamics-based world model to predict the resulting future visual features, while the progress predictor assigns a differentiable score to the predicted outcome. We then backpropagate the progress gradients through the world model to provide classifier-style guidance during each diffusion denoising step. This steers the sampling process toward action chunks predicted to maximize progress toward the goal. By coupling planning with evaluation, progress shapes the generated trajectories and provides a simple threshold-based termination criterion, leading to more goal-directed and reliable long-horizon execution.

In summary, our contributions are threefold: (1) a progress estimator grounded in predicted future observations through an inverse dynamics world model, enabling foresight in task evaluation; (2) a progress-guided diffusion sampler that leverages differentiable progress gradients to iteratively optimize action chunks during generation; and (3) extensive empirical validation on the CALVIN and LIBERO benchmarks, complemented by real-robot deployments, demonstrating significant gains in long-horizon success rates and more reliable task termination.

\begin{figure*}[t]
\centering
\includegraphics[width=0.9\textwidth]{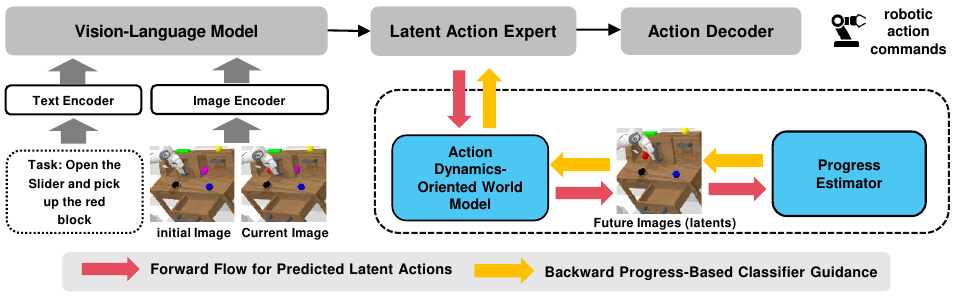}
\caption{Overview of \vla. Conditioned on a language instruction and current observation, the diffusion policy first generates a candidate chunk of latent actions. A action-oriented world model then rolls out these actions within a pre-trained visual feature space to project future states, while a progress estimator assigns a completion score to the predicted outcomes. Finally, progress gradients are backpropagated through the world model as classifier guidance, steering the diffusion process toward actions that maximize task advancement.}
\label{fig:overview}
\vspace{-0.1in}
\end{figure*}

\section{Related Work}

\subsection{Vision-Language-Action Models}

Large-scale Vision-Language Models (VLMs) have established robust multimodal representations that transfer effectively across diverse perception tasks, including visual question answering and image captioning. Extending these capabilities to control settings has led to the emergence of Vision-Language-Action (VLA) models~\cite{ma2024survey,sapkota2025vision}, which map linguistic goals and visual observations to control policies. Current VLA research generally follows three paradigms: (i) autoregressive tokenization, which discretizes continuous control signals into action codebooks~\cite{kim2024openvla,pertsch2025fast,bu2025univla}; (ii) direct supervised regression to joint action spaces when dense labels are available~\cite{zhao2023learning,kim2025fine,wang2025vla}; and (iii) generative trajectory modeling, notably diffusion policies that sample action sequences via iterative denoising~\cite{chi2025diffusion,ke20243d,black2024pi_0,intelligence2504pi0,bu2025agibot}. While diffusion-based methods produce high-quality, diverse trajectories through stochastic sampling, they often lack an explicit task-level signal to steer generation toward goal completion. Parallel efforts have focused on learning latent, transferable action spaces from video data to facilitate cross-embodiment generalization~\cite{ye2024latent,bruce2024genie,bu2025laof,bi2025motus,chen2024igor,chen2025villa,chen2024moto,li2025latbot,bu2025agibot}. These approaches typically utilize an Inverse Dynamics Model (IDM) to infer latent actions from video frames and a Forward Dynamics Model (FDM) to reconstruct future states. For instance, recent works~\cite{ye2024latent,bu2025univla,chen2025villa,bu2025agibot} demonstrate that learning compact, latent action representations from large-scale human data can mitigate domain gaps and provide a powerful supervisory signal for next-token prediction, ultimately yielding higher-fidelity robotic trajectories.


Despite recent advancements, existing methods primarily rely on passive vision-language conditioning and fail to incorporate an explicit mechanism for monitoring task progress or completion. In contrast, we introduce a progress-critic framework that integrates a learned progress estimator directly into diffusion-based sampling within a latent action space. By fusing transferable latent representations with progress-guided generative planning, our approach facilitates more goal-directed and robust trajectories, significantly enhancing performance in challenging, long-horizon manipulation tasks.

\subsection{World Model}

World models establish compact internal representations of environmental dynamics, facilitating prediction, planning, and counterfactual reasoning without the need for costly physical rollouts~\cite{li2025comprehensive}. In robotics, these models are increasingly utilized to learn latent dynamics for model-based control, synthesize future observations for imagination-based planning~\cite{team2025gigaworld,jiang2025enerverse,hung2025nora,cen2025rynnvla}, and estimate task-specific objectives such as success classifiers or reward functions~\cite{xiao2025world,zhu2025wmpo}. While such signals enable reinforcement learning through simulated rollouts, the resulting supervision is often restricted to sparse, binary success indicators that provide a limited gradient for efficient optimization~\cite{fei2025srpo}. Recent advancements~\cite{song2025reconvla,bu2025agibot} address this by demonstrating that the joint learning of latent dynamics alongside perception and action embeddings significantly enhances sample efficiency and cross-embodiment generalization.


In this paper, we introduce a progress-oriented model that explicitly predicts a scalar progress estimate alongside latent observation dynamics. This learned signal serves as a dense, task-oriented guidance mechanism during diffusion-based sampling and can be thresholded to establish reliable, principled termination criteria. By jointly modeling latent actions, state dynamics, and task progress, our world model facilitates highly goal-directed trajectory generation while significantly reducing dependence on costly, sparse, or task-specific supervision.


\section{The Proposed Method}
\label{sec:method}


Given an image observation $o$ and a language instruction $l$, our goal is to train a policy $\pi$ to predict a coherent action chunk, denoted as $\pi: (o_t, l) \rightarrow a_{t:t+N}$. Our proposed \vla~framework achieves this through three components: (1) a progress estimator that regresses a normalized task-completion score from the current visual and language instruction (Sec.~\ref{sec:method-progress}); (2) an action-conditioned world model that facilitates bidirectional reasoning by either projecting future visual states from predicted latent actions (forward dynamics) or inferring the underlying latent actions from visual state transitions (inverse dynamics). (Sec.~\ref{sec:method-world}); and (3) a diffusion-based generative model that leverages the progress signal through differentiable classifier guidance to steer action sampling toward goal-optimal trajectories (Sec.~\ref{sec:method-diffusion}). See Fig.~\ref{fig:overview} for an overview.


\subsection{Progress Estimator}
\label{sec:method-progress}


The progress estimator $P$ operates as a vision-language evaluator that assesses task advancement. It processes the language instruction $l$, the initial observation $o_0$ (to provide a global task anchor), and the current image $o_t$ to output a normalized scalar progress score:
\begin{equation}
\label{eq:progress_pred}
 p = P(l, o_0, o_t), \  p\in[0,1].
\end{equation}
We train $P$ as a regressor with an $L_1$ loss:
\begin{equation}
\label{eq:progress_loss}
\mathcal{L}_{\text{prog}} = | p - p^* |,
\end{equation}
where $p^*$ denotes the ground-truth progress label.


We utilize the normalized timestep as a proxy for progress; specifically, for a trajectory of total length $T$, the progress label at timestep $t$ is defined as $p^* = t / T$. This formulation is grounded in the observation that our expert demonstrations are curated to advance steadily toward completion, ensuring that task progress remains approximately monotonic. Consequently, normalized time serves as a robust and effective surrogate for progress, without requiring additional annotations. 







\begin{figure}[t]
\centering
\setlength{\fboxsep}{6pt}
\includegraphics[width=0.95\linewidth]{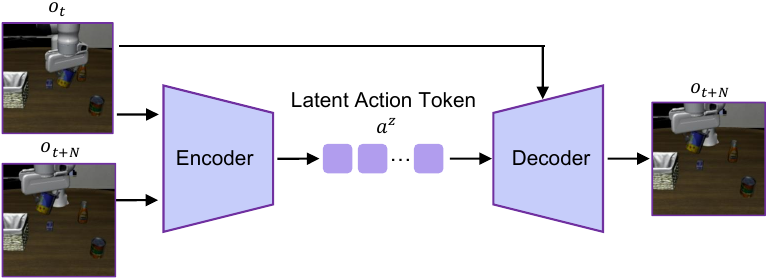}
\caption{Architecture of action dynamics oriented world model.}
\vspace{-0.1in}
\label{fig:wm}
\end{figure}

\subsection{World Model}
\label{sec:method-world}

Our method incorporates a compact latent world model designed to capture visual states and dynamics for short-horizon imagination. This architecture as in Fig.~\ref{fig:wm} consists of a vision encoder $E$ and a decoder $D$. Specifically, the encoder serves as an inverse dynamics model, mapping the transition between two observations, $o_t$ and $o_{t+N}$, into a compressed latent action space:
\begin{equation}
\label{eq:latent_action}
a^z = E(o_t, o_{t+N}),\\
\end{equation}
and the decoder (the forward dynamics model) predicts the future image given observation $o_t$ and latent action $a^z$:
\begin{equation}
\label{eq:latent_loss}
o_{t+N} = D(o_t, a^z).
\end{equation}
The training objective for the proposed world model integrates a latent-dynamics reconstruction loss with a Kullback-Leibler (KL) divergence term to regularize the latent action distribution ($\mathcal{N}$ is the normal distribution):
\begin{equation}
\label{eq:world_loss}
\mathcal{L}_{\text{world}} = \sum_t \|o_{t+N}-o^{*}_{t+N}\|^2 + KL(a^z, \mathcal{N}(0,I)).
\end{equation}



Essentially, we train the world model to extract compact, transferable latent representations that decouple visual nuisances from task-relevant features. These latents serve as a unified state representation, shared by both the policy generator for actions and the progress estimator for state evaluation.


\subsection{Joint Finetuning of World Model and Progress Estimator}


After the separate pre-training of the world model and the progress estimator, we perform joint fine-tuning to align latent dynamics with task-level progression. Specifically, given a current visual observation $o_t$ and a candidate latent-action chunk $a^z_{t:t+N}$, the world model first projects the resulting future latent state; the progress estimator then assesses this predicted outcome to compute a task-advancement score:
\begin{equation}
 p_{t+N} = P(l, o_0, D(o_t, a^z_{t:t+N})).
\end{equation}
We define a loss on the predicted progress score, which jointly supervises the two modules: 
\begin{equation}
\mathcal{L}_{\text{joint}} = \| p_{t+N} - p^*_{t+N} \|.
\end{equation}
The overall joint finetuning objective is a weighted average of the world-model loss, progress loss and joint loss, namely
\begin{equation}
\mathcal{L}_{\text{ft}} = \mathcal{L}_{\text{world}} +\mathcal{L}_{\text{prog}}+\mathcal{L}_{\text{joint}},
\end{equation}
which encourages the predicted latent dynamics to be informative for downstream progress estimation and guidance.


\subsection{Progress-Guided Diffusion Policy}
\label{sec:method-diffusion}


Our policy employs a two-stage generation pipeline designed for cross-embodiment flexibility. First, a Latent Action Expert generates an action chunk $a^z_{t:t+N}$ within an embodiment-agnostic latent space, focusing on high-level task strategy. In the second stage, an Action Decoder maps $a^z_{t:t+N}$ into a low-level action sequence $a_{t:t+N}$ for robot execution. 


Let $x_0$ denote the latent representation of $a^z$ as in Eq.~\ref{eq:latent_action}. The backbone diffusion model is trained by optimizing the denoising objective, namely
$$\mathcal{L}_{\text{diff}} = \mathbb{E}_{x_0,\epsilon,\tau}\big\|\epsilon - \epsilon_{\theta}(x_{\tau},\tau,l,o_t)\big\|^2,$$
where $x_{\tau}$ is the noisy latent-action sample at diffusion step $\tau$ (distinct from the observation $o_t$), $\epsilon$ is a Gaussian noise, and $\epsilon_{\theta}$ predicts the noise.

The diffusion policy is guided with the progress estimator through the world model. Given the current visual observation $o_t$ and the current latent sample $x_{\tau}$, the world model predicts the resultant future image:
\begin{equation}
\hat{o}_{t+N} = D(o_t, x_{\tau}),
\end{equation}
which is then fed to the progress estimator to obtain the predicted progress via:
\begin{equation}
\hat p_{t+N} = P(l, z_0, \hat{o}_{t+N}).
\end{equation}
Since $\hat p_{t+N}$ is differentiable with respect to $\mathrm{x}_{\tau}$ through the world model decoder $D$, we can backpropagate gradients to the latent action and use them as classifier guidance during sampling. At diffusion step $\tau$, let the unguided reverse mean be $\mu_\theta(x_{\tau},\tau,c)$. We modify the update rule as
\begin{equation}
x_{\tau-1} = \mu_{\theta}(x_{\tau},\tau,c) + s\,\nabla_{x_{\tau}}\hat p_{t+N} + \sigma_{\tau}\epsilon,
\end{equation}
where $s$ controls the guidance strength and $\nabla_{x_{\tau}}\hat p_{t+N}$ effectively optimizes $a^z$ towards increased progress.


The sampled latent-action chunk $x_0$ is mapped by the action decoder into an executable sequence $a_{t:t+N-1}$ for deployment. Empirical results demonstrate that progress-guided sampling significantly shifts the distribution of generated actions toward those yielding higher predicted progress; this effectively reduces the need for extensive re-sampling and enables a robust, threshold-based termination criterion at runtime.

\begin{figure}[t]
\centering
\setlength{\fboxsep}{6pt}
\includegraphics[width=0.98\linewidth]{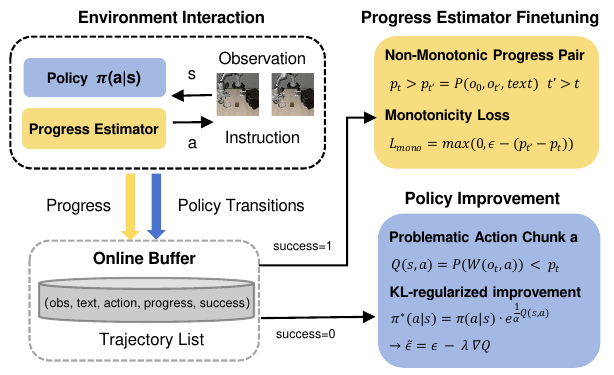}
\caption{Reinforcement learning framework of \vla.}
\label{fig:rl}
\vspace{-0.1in}
\end{figure}

\subsection{RL Finetuning with Progress}
\label{sec:method-rl}

While progress-guided sampling can be applied at inference time, we further \emph{finetune} both the progress estimator and the diffusion policy with online experience (see Fig.~\ref{fig:rl}), so that (i) progress estimates remain well-aligned with task completion and (ii) the resulting policy is more robust to execution noise.


\paragraph{Online trajectory collection}
We periodically roll out the current policy to collect trajectories
\begin{equation}
\mathcal{B} = \{(o_0, o_t, l, a^z_{t:t+N}, a_{t:t+N}, \hat p_t, y)\}_{t=0}^{T-1},
\end{equation}
where $\hat p_t$ is the predicted progress, and $y\in\{0,1\}$ indicates episode success. This online buffer captures critical edge cases that are typically under-represented in static offline datasets, such as recovery behaviors, near-failure states, and out-of-distribution visual perturbations.


\paragraph{Progress estimator finetuning}
For successful episodes ($y{=}1$), task progress should be (approximately) monotonic. We therefore mine progress anomalies, namely instances where the predictor's output violates this expected monotonicity. Specifically, for each timestep $t$ we define
\begin{equation}
t' = \arg\min_{k>t}\ \hat p_k,
\end{equation}
namely the index after $t$ with the smallest predicted progress, and mark $t$ as an anomaly if the following holds:
\begin{equation}
\mathcal{I}_{\text{anom}}=\{t \mid \hat p_t > \hat p_{t'} + \epsilon\}.
\end{equation}
and apply a margin-based monotonicity loss
\begin{equation}
\mathcal{L}_{\text{mono}}=\sum_{t\in \mathcal{I}_{\text{anom}}}\max\big(0,\ \epsilon-(\hat p_{t'}-\hat p_t)\big).
\end{equation}

In implementation, we finetune $P$ by minimizing $\mathcal{L}_{\text{prog}}+\mathcal{L}_{\text{mono}}$ on the online buffer.

\paragraph{Diffusion policy finetuning}
We cast progress maximization as a KL-regularized policy improvement. Let the state be $s=(l,o_0,o_t)$ and the action be the latent action $a^z$. We define a task-aware score using the learned evaluator:
\begin{equation}
Q(s,a)=P(l,o_0,\ D(o_t,a^z)),
\end{equation}
\emph{i.e.}, the progress predicted after applying $a$ to the world model. We then formulate the following KL-constrained optimization problem:
\begin{equation}
\begin{gathered}
\pi^{*}(a|s)\ =\ \text{argmax}_{\pi^*_{\theta}(a|s)}E_{a\sim\pi^*_{\theta}(\cdot|s)}\big[Q(s,a)\big],\\
\text{s.t.}\quad \mathrm{KL}(\pi^{*}(\cdot|s)\,\|\,\pi_\theta(\cdot|s)) \le \varepsilon.
\end{gathered}
\end{equation}
Solving this problem yields 
\begin{equation}
\pi^{*}(a|s)\ \propto\ \pi_{\theta}(a|s)\exp\!\left(\tfrac{1}{\alpha}Q(s,a)\right),
\end{equation}
which increases the likelihood of action chunks linked with higher progress while staying close to the current policy.

For diffusion policies, $\pi_\theta$ is parameterized by the denoiser $\epsilon_\theta$. From a guided denoising view, we incorporate the progress score by adjusting the noise target at each diffusion step:
\begin{equation}
\tilde\epsilon=\epsilon - \tfrac{\sigma_t}{\alpha}\nabla_{x_{\tau}}Q(s,x_{\tau}),
\end{equation}
and train the policy with a standard denoising objective,
\begin{equation}
\mathcal{L}_{\text{policy}}=\mathbb{E}\big[\ \|\tilde\epsilon-\epsilon_\theta(x_{\tau},\tau,l,o_t)\|^2\ \big].
\end{equation}
This update encourages the denoiser to produce samples that move toward higher progress.

\begin{table*}[t]
  \centering
  \caption{The comparisons with state-of-the-art approaches on CALVIN (ABC$\rightarrow$D) with the metrics of success rate and average success
  length. The abbreviations denote different input modalities: S-RGB for Static RGB, G-RGB for Gripper RGB, S-RGBD for Static RGB-D, G-RGBD for Gripper RGB-D, P for proprioceptive arm position, and Cam for camera parameters.}
  \label{tab:calvin2}
  \resizebox{0.9\linewidth}{!}{%
  \begin{tabular}{llcccccc}
  \toprule
  Method & Input & \multicolumn{5}{c}{Task completed in a Row(\%) $\uparrow$} & Avg.Len. \\
  \cmidrule(lr){3-7}
   & & 1 & 2 & 3 & 4 & 5 & \\
  \midrule
  RoboFlamingo \cite{li2023vision} & S-RGB,G-RGB & 82.4 & 61.9 & 46.6 & 33.1 & 23.5 & 2.47 \\
  GR-1 \cite{wu2023unleashing} & S-RGB,G-RGB,P & 85.4 & 71.2 & 59.6 & 49.7 & 40.1 & 3.06 \\
  3D Diffuser \cite{ke20243d} & S-RGBD,G-RGBD,P,Cam & 92.2 & 78.7 & 63.9 & 51.2 & 41.2 & 3.27 \\
  GR-MG \cite{li2025gr} & S-RGBD,G-RGBD,P & {96.8} & {89.3} & {81.5} & {72.7} & 64.4 & {4.04} \\
  \midrule
  SuSIE \cite{black2023zero} & S-RGB & 87.0 & 69.0 & 49.0 & 38.0 & 26.0 & 2.69 \\
  GHIL-Glue \cite{black2023zero, hatch2025ghil} & S-RGB & 95.2 & \textbf{88.5} & 73.2 & 62.5 & 49.8 & 3.69 \\
  Dita \cite{hou2025dita} & S-RGB & 94.5 & 82.5 & 72.8 & 61.3 & 50.0 & 3.61 \\
  \midrule
  \textbf{\vla~(w/o cg)} & S-RGB & 89.4 & 76.8 & 63.0 & 52.2 & 43.1 & 3.24 \\
  \textbf{Pretrained \vla~(w/o cg)} & S-RGB & 92.7 & 81.6 & 70.1 & 60.9 & 51.6 & 3.57 \\
  \textbf{Pretrained \vla~(w/ cg)} & S-RGB & 93.6 & 82.4 & 71.2 & 60.8 & 52.8 & 3.61 \\
  \textbf{Pretrained \vla~(w/ cg(pretrained))} & S-RGB & 93.6 & 82.0 & 72.0 & 63.6 & \textbf{56.4} & 3.68 \\
  \textbf{Pretrained \vla~(Full)} & S-RGB & \textbf{95.2} & 84.8 & \textbf{73.6} & \textbf{67.2} & 52.0 & \textbf{3.73} \\
  \bottomrule
  \end{tabular}
  }
  \end{table*}

\section{Experiments}
\subsection{Pretraining Data}

All components are pre-trained on the Open X-Embodiment (OXE) datasets~\cite{o2024open, kim2024openvla}, adhering to the dataset selection and mixture weighting protocols established in~\cite{kim2024openvla, team2024octo}. Actions are normalized and filtered using the same procedure as~\cite{o2024open}. Unless otherwise specified, we use the same image preprocessing across all modules. The modules are trained with a batch size of 2048 on 8 NVIDIA H20 GPUs (256 samples per GPU) and the base learning rate is set to $1\times10^{-4}$.

\subsection{Implementation Details}
\subsubsection{Progress estimator pretraining}

In implementation, the progress estimator takes the patch features of the starting and current frames extracted by DINOv2~\cite{oquab2023dinov2} as input. Visual and text tokens are first projected into a shared embedding space, where learnable role embeddings (start, current, and text) preserve functional distinctness. A lightweight cross-attention stack, with residual connections and LayerNorm, aligns the language instructions with the current observation while encoding start-to-current changes. Finally, the tokens are mean-pooled and fused via an MLP, with a sigmoid head outputting the scalar progress score $p$.

\subsubsection{World model pretraining}
We adopt the UniVLA~\cite{bu2025univla} world model architecture to predict future visual features given candidate latent actions. To stabilize downstream latent-action prediction, we add a KL regularization term during training to normalize the latent action distribution (Eq.~\ref{eq:world_loss}), which improves compatibility with the Latent Action Expert.

\subsubsection{Latent action expert pretraining}
Our Latent Action Expert follows the DiTA-style design~\cite{hou2025dita} and uses a causal Transformer to autoregressively predict latent actions from multimodal context. The Action Decoder shares the same architecture and training recipe. Additional hyperparameters are deferred to the supplemental material.

\subsection{CALVIN}
\label{sec:calvin}

CALVIN~\citep{mees2022calvin} is a simulated benchmark for long-horizon, language-conditioned manipulation. It contains four distinct environments (A, B, C, and D). We adopt the standard ABC$\rightarrow$D evaluation protocol, training on environments A, B, and C while testing on D. Each evaluation trial involves a sequence of five subtasks sampled from a diverse pool of language-specified goals, totaling up to 1,000 unique sequences. Following established metrics, we report the success sequence length (denoted as ``task completed in a row''), namely consecutive subtasks completed in a row, from 1 to 5, alongside the average number of tasks completed per episode.

\subsubsection{Baselines and our variants}
The adopted competing methods for comparison can be found in  Table~\ref{tab:calvin2}. In addition, we consider the following variants:
\begin{itemize}
  \item \textbf{\vla~(w/o CG).} From-scratch trained diffusion policy without classifier (progress estimator) guidance.
  \item \textbf{Pretrained \vla~(w/o CG).} Pretrained diffusion policy on OXE with no guidance at inference.
  \item \textbf{Pretrained \vla~(w/ CG).} Pretrained diffusion policy with classifier guidance, where evaluator trained from scratch on CALVIN.
  \item \textbf{Pretrained \vla~(w/ Pretrained CG).} Pretrained diffusion policy with classifier guidance, using the pretrained world model and progress predictor as evaluator.
  \item \textbf{Pretrained \vla~(Full).} Pretrained diffusion policy with classifier guidance using the pretrained evaluator + RL finetuning.
\end{itemize}

\subsubsection{Pretraining contributes to diffusion policy performance}
Comparing {Pretrained \vla~(w/o CG)} to {\vla~(w/o CG)} in Table~\ref{tab:calvin2} shows that pretraining the diffusion policy yields a large and consistent improvement in overall task completion performance, especially on longer-horizon sequences. 
This indicates that DP pretraining provides a strong latent-action prior and  reduces compounding errors even without  guidance.

\subsubsection{Classifier guidance relies on a reliable evaluator} Adding guidance on top of a pretrained policy ({Pretrained \vla~(w/o CG)} $\rightarrow$ {Pretrained \vla~(w/ CG)}) yields a moderate improvement. Notably, the benefit of classifier guidance becomes significantly larger when the evaluator (world model + progress predictor) is pretrained: {Pretrained \vla~(w/ CG)} $\rightarrow$ {Pretrained \vla~(w/ CG (pretrained))} increases the 5-in-a-row rate from $52.8\%$ to $56.4\%$ (+3.6) and the 4-in-a-row rate from $60.8\%$ to $63.6\%$ (+2.8), while improving the average completed length to $3.68$. We attribute this gap to the \emph{robustness} of the pretrained vision-language evaluator: pretraining yields a more reliable progress signal under distribution shift and execution noise and provides higher-quality guidance gradients during sampling.

\subsubsection{RL finetuning further improves robustness}
\label{sec:rl_finetune}

In the RL finetuning stage, we roll out the  policy in training environments A/B/C and collect a total of 1,000 trajectories for online updates. {Pretrained \vla~(Full)} achieves the best overall average completed length and improves 1--4 subtask success.
We attribute these gains to the complementary effects of RL: online experience improves progress--completion alignment in the evaluator and yields a stronger guidance signal. And it simultaneously refines the diffusion policy to be more robust to execution noise.

\subsection{LIBERO}
  \label{sec:libero}
  
LIBERO~\citep{liu2023libero} is a comprehensive benchmark for knowledge transfer in multitask and lifelong robot learning. It contains four sub-datasets: LIBERO-SPATIAL, LIBERO-OBJECT, LIBERO-GOAL, and LIBERO-100. LIBERO-100 is further split into LIBERO-90 and LIBERO-LONG, where LIBERO-LONG features long-horizon tasks that require diverse object interactions and versatile motor skills. We use the modified LIBERO setup released with OpenVLA~\cite{kim2024openvla} as the data source for finetuning and evaluation. 
  
\begin{table}[t]
\centering
\caption{The experimental results on the LIBERO benchmark. See the main text for more explanation.}
\label{tab:table_libero}
\small
\setlength{\tabcolsep}{5pt}
\begin{tabular}{lccccc}
\toprule
Method & Spatial & Object & Goal & Long & Average \\
\midrule
LAPA~\citep{ye2024latent} & 73.8 & 74.6 & 58.8 & 55.4 & 65.7 \\
Diffusion Policy~\citep{chi2025diffusion} & 78.3 & 92.5 & 68.3 & 50.5 & 72.4 \\
Octo~\citep{team2024octo} & 78.9 & 85.7 & 84.6 & 51.1 & 75.1 \\
MDT~\citep{reuss2024multimodal} & 78.5 & 87.5 & 73.5 & 64.8 & 76.1 \\
OpenVLA~\citep{kim2024openvla} & 84.7 & 88.4 & 79.2 & 53.7 & 76.5 \\
MaIL~\citep{jia2024mail} & 74.3 & 90.1 & 81.8 & \textbf{78.6} & 83.5 \\
Dita~\cite{hou2025dita} & 84.2 & 96.3 & 85.4 & 63.8 & 82.4 \\
\midrule
\textbf{\vla~w/o cg} & 83.2 & 95.0 & 84.6 & 63.2 & 81.5 \\
\textbf{\vla~w/ cg} & 85.8 & 96.1 & 86.0 & 65.4 & 83.3 \\
\textbf{\vla~Full} & \textbf{88.2} & \textbf{96.4} & \textbf{87.2} & 66.2 & \textbf{84.5} \\
\bottomrule
\end{tabular}
\end{table}

Table~\ref{tab:table_libero} reports success rates (\%) on LIBERO-SPATIAL, LIBERO-OBJECT, LIBERO-GOAL, and LIBERO-LONG, together with the average across subsets. We compare against representative multitask VLA baselines (\emph{e.g.}, Diffusion Policy~\citep{chi2025diffusion}, OpenVLA~\citep{kim2024openvla}, MDT~\citep{reuss2024multimodal}, and Dita~\cite{hou2025dita}). To isolate the contribution of progress guidance, we report three variants: \textbf{w/o cg} performs unguided diffusion sampling (no classifier guidance); \textbf{w/ cg} enables progress-guided classifier guidance at inference; and \textbf{Full} denotes our strongest configuration. Across all subsets, progress guidance yields consistent gains over the unguided counterpart (\emph{e.g.}, average $81.5\!\rightarrow\!83.3$ and LIBERO-LONG $63.2\!\rightarrow\!65.4$), while the full model further improves performance (average $84.5$). Notably, our full method achieves the best overall average and delivers strong improvements on the long-horizon LIBERO-LONG split compared to OpenVLA ($66.2$ vs.\ $53.7$), supporting the effectiveness of progress-guided diffusion policy.

\begin{figure*}[h]
  \centering
  \includegraphics[width=0.9\textwidth]{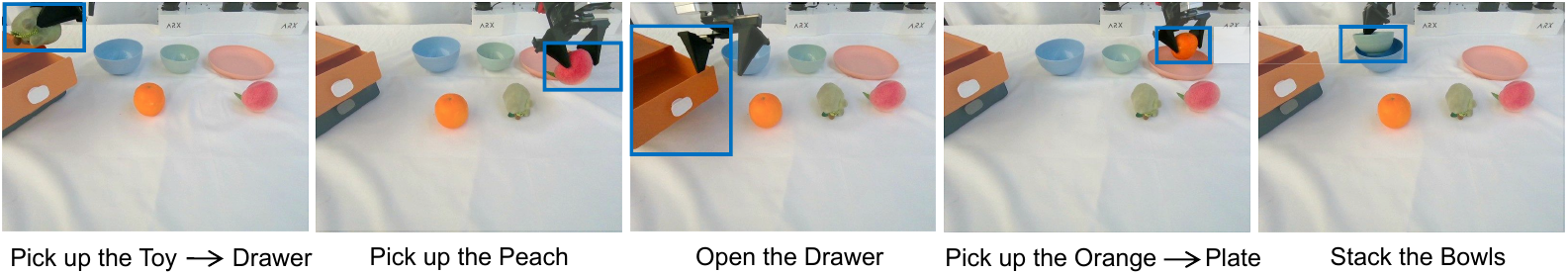}%
  \caption{Illustration of the five tasks in real-world model deployment on an ARX robotic dual-arm.}
  \label{fig:real_world_exp}
\end{figure*}

\begin{table*}[t]
  \centering
  \caption{\textbf{Real-robot results on five tasks.} We report average end-effector path length (Avg. distance, m), success rate (\%), and average steps (Avg. steps).}
  \label{tab:real_robot_5tasks}
  \resizebox{0.9\linewidth}{!}{%
  \begin{tabular}{lccc|ccc|ccc}
  \toprule
  & \multicolumn{3}{c|}{Octo\citep{team2024octo}} & \multicolumn{3}{c|}{ProgressVLA (w/o CG)} & \multicolumn{3}{c}{ProgressVLA (w/ CG)} \\
  \cmidrule(lr){2-4}\cmidrule(lr){5-7}\cmidrule(lr){8-10}
  Task & Success$\uparrow$ & Dist$\downarrow$ &  Steps$\downarrow$
       & Success$\uparrow$ & Dist$\downarrow$ &  Steps$\downarrow$
       & Success$\uparrow$ & Dist$\downarrow$ &  Steps$\downarrow$ \\
  \midrule
  Pick up the peach
    & 35 & 1.04 & 151.6 & 70 & 0.78 & 96.0 & 90 & 0.61 & 39.1 \\
  Open the drawer
    & 20 & 1.45 & 196.8 & 80 & 0.92 & 90.5 & 90 & 0.78 & 42.0 \\
  Stack the bowls
    & 30 & 0.91 & 145.8 & 65 & 0.71 & 94.1  & 70 & 0.59 & 37.8 \\
  Pick up the orange $\rightarrow$ plate
    & 20 & 1.63 & 228.3 & 60 & 1.38 &  117.2  & 75 & 1.12 & 72.0 \\
  Pick up the toy $\rightarrow$ drawer
    & 10 & 1.48 & 216.8 & 55 & 1.03 & 106.1  & 55 & 0.96 & 75.6 \\
  \midrule
  Avg.
    & 23 & 1.30 & 187.9 & 66 & 0.96 & 100.8  & {76} & {0.81} & {53.3} \\
  \bottomrule
  \end{tabular}
  }
\end{table*}

\subsection{Real-World Evaluation}
\label{sec:real_world}

\subsubsection{Experiment setup.}

Real-world experiments are conducted using an ARX AC-One robotic dual-arm outfitted with two X5 arms and ARX G2 parallel grippers. The sensory setup consists of two Intel RealSense D405 RGB-D cameras: one wrist-mounted and one positioned as a stationary third-person "head" view. While the cameras support depth, we utilize RGB images as the primary policy inputs unless specified otherwise. All trials are performed in a tabletop manipulation environment characterized by fixed object initializations and consistent camera perspectives.

\subsubsection{Task setup.}
We evaluate the proposed model on five real-robot manipulation tasks of varying difficulty, as shown in Fig.~\ref{fig:real_world_exp}: {Toy$\rightarrow$Drawer}, {Pick Peach}, {Open Drawer}, {Orange$\rightarrow$Plate}, and {Stack Bowls}. These tasks span both single-step and long-horizon behaviors. Target objects and goal receptacles are highlighted with blue boxes in Fig.~\ref{fig:real_world_exp}.

\subsubsection{Data collection and finetuning.}
For each task, we collect 50--100 human teleoperated trajectories, depending on task complexity, to finetune the models. Each trajectory contains multi-view RGB observations from the wrist and head cameras together with the executed action sequence. 

\subsubsection{Evaluation protocol and baselines.}
For quantitative evaluation, we run 20 trials per task. We compare against two baselines: (i) Octo~\citep{team2024octo}, a strong pretrained VLA policy, and (ii) \vla~(w/o CG), which serves as an unguided generative baseline. Our full method applies progress-guided classifier guidance during sampling to improve task completion reliability.

Table~\ref{tab:real_robot_5tasks} reports real-robot results using success rate ({Succ}, \%), end-effector travel distance ({Dist}, m), and executed action chunks ({Steps}; one step corresponds to one predicted and executed chunk). Lower Dist/Steps indicates more efficient execution. Overall, \vla~substantially outperforms Octo, and classifier guidance (CG) further improves both reliability and efficiency. Averaged across tasks, Octo achieves {23\%} success with {1.30\,m}/{187.9} steps, while {\vla~(w/o CG)} increases success to {66\%} and reduces Dist/Steps to {0.96\,m}/{100.8}. With CG, {\vla~(w/ CG)} further improves to {76\%} success with {0.81\,m}/{53.3} steps. The gains are especially clear on tasks that otherwise exhibit redundant motion, suggesting that progress-guided CG leads to more decisive and goal-directed real-robot execution.

\begin{figure}[b]
    \centering
    \includegraphics[width=\linewidth]{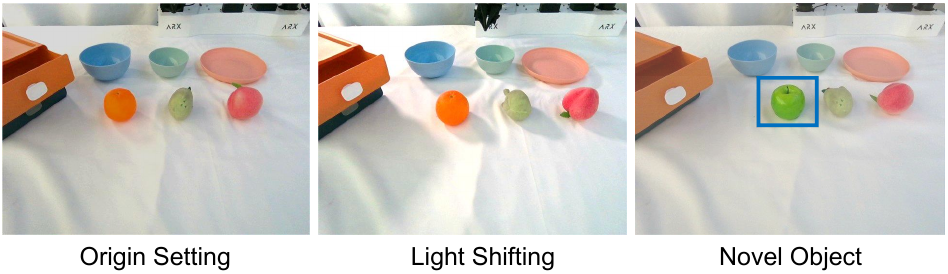}%
    \caption{Real-world scenarios used for investigating the generalization of progress estimator.}
    \label{fig:real_world_scenes}
\end{figure}

\subsection{Progress Estimator Generalization on Real-Robot Expert Trajectories}
  \label{sec:pp_generalization_real}

\subsubsection{Offline, policy-agnostic evaluation protocol.}
We evaluate the progress predictor independently of policy learning using a small set of real-robot expert demonstrations collected under three controlled scene settings, as shown in Fig.~\ref{fig:real_world_scenes}: {(i) Original}, {(ii) Lighting shift} (adding a desk lamp), and {(iii) Novel objects} (swapping the target object with an unseen instance while keeping the layout and cameras fixed).

Given the language instruction $\ell$ and observation $o_t$ (optionally with $o_0$), the progress estimator outputs a normalized progress $\hat p_t\in[0,1]$ at each timestep.

\begin{figure*}[t]
  \centering
\includegraphics[width=0.9\linewidth]{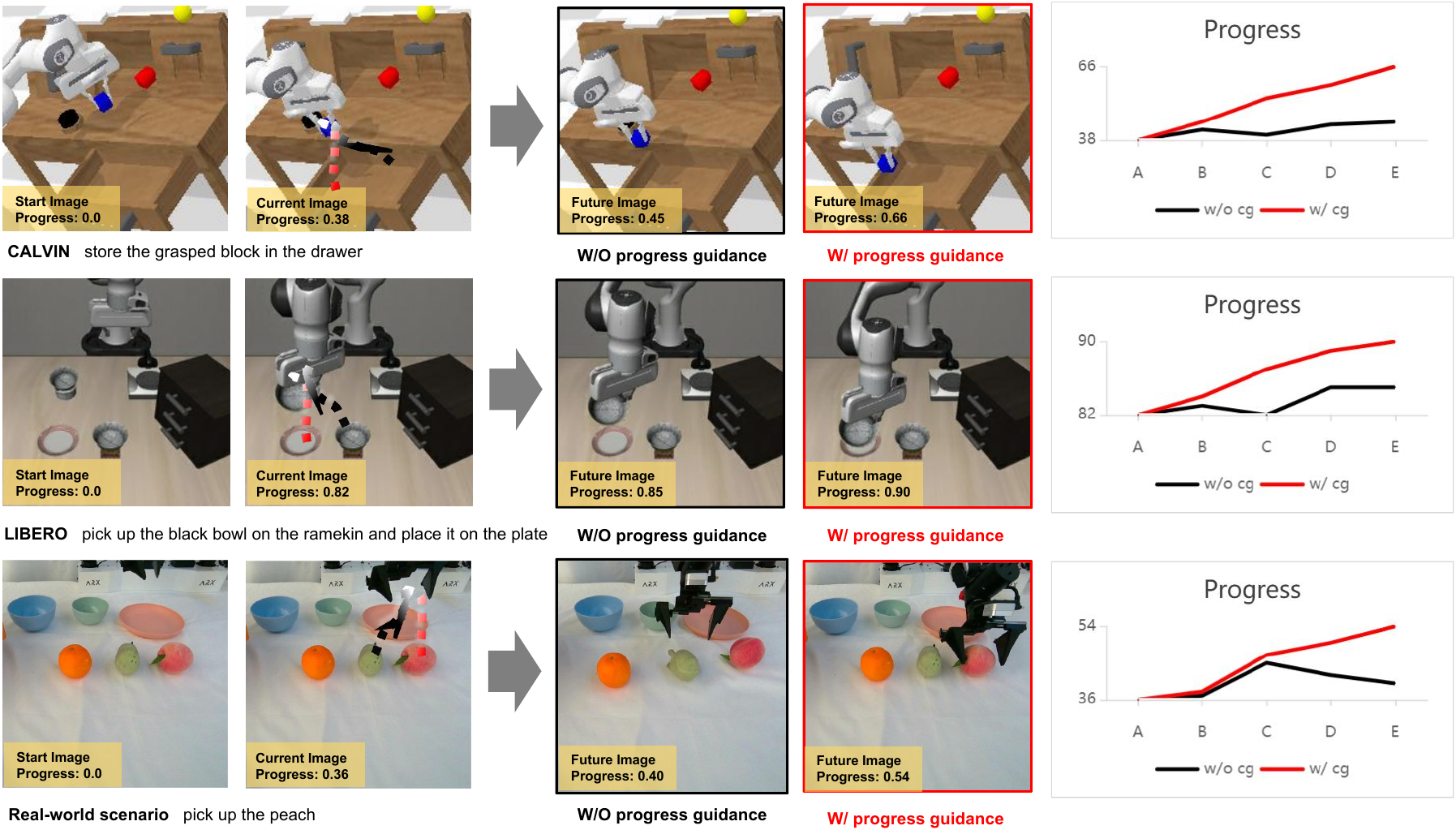}
  \caption{Visualization of progress estimation and effect of guidance. The second column from the left illustrates the predicted trajectories, contrasting the baseline diffusion policy (in black) with our \vla~approach (in red). Our method consistently generates more plausible, goal-directed paths.}
  \label{fig:visualization}
  
\end{figure*}

\begin{table}[h]
\centering
\caption{``From Scratch" trains the progress predictor from random initialization on our real-robot data, while ``Finetuned" starts from the pretrained checkpoint and is finetuned on the same data.}
\small
\setlength{\tabcolsep}{6pt}
\label{tab:real_world_scenes}
\begin{tabular}{llccc}
\toprule
Setting & Method & Pearson$\uparrow$ & Stop$\uparrow$ & MAE$\downarrow$ \\
\midrule
\multirow{2}{*}{Origin Setting}
  & From Scratch & 0.912 & 53.8 & 0.14 \\
  & Finetuned    & 0.977 & 82.1 & 0.10 \\
\midrule
\multirow{2}{*}{Lighting Shift}
  & From Scratch & 0.809 & 3.6  & 0.24 \\
  & Finetuned    & 0.953 & 80.8 & 0.12 \\
\midrule
\multirow{2}{*}{Novel Objects}
  & From Scratch & 0.810 & 37.5 & 0.15 \\
  & Finetuned    & 0.972 & 81.2 & 0.11 \\
\bottomrule
\end{tabular}

\end{table}

\subsubsection{Metrics}
We report three metrics, as seen in Table~\ref{tab:real_world_scenes}: {(i) Progress alignment} (Pearson correlation between $\{\hat p_t\}$ and a reference ramp $p_t=t/T$; higher is better), {(ii) Stop reliability} (fraction of trajectories with $\max_{t\in\{T-9,\ldots,T\}}\hat p_t>0.9$; higher is better), and {(iii) Progress error} (MAE between $\hat p_t$ and $p_t$; lower is better).

\subsubsection{From-scratch vs.\ pretrained and finetuned progress predictor}
  We compare two training regimes:
  \emph{From Scratch}, where the progress predictor is trained only on the target real-robot dataset without large-scale pretraining, and
  \emph{Finetuned}, where we start from a progress predictor pretrained on large-scale manipulation data and then finetune on the real-robot demonstrations.

\subsubsection{Results and discussion}
Table~\ref{tab:real_world_scenes} shows that {pretraining} is critical for progress generalization, and the pretrained model {remains robust after finetuning} on real-robot data. In the {original} scene, the pretrained+finetuned predictor outperforms training from scratch across all metrics (Pearson $0.912\!\rightarrow\!0.977$, Stop $53.8\!\rightarrow\!82.1$, MAE $0.14\!\rightarrow\!0.10$). Under {lighting shift}, the from-scratch model degrades sharply (Pearson $0.809$, Stop $3.6$), whereas the pretrained+finetuned model stays strong (Pearson $0.953$, Stop $80.8$, MAE $0.12$). A similar trend holds for {novel objects}, where pretraining yields large gains (Pearson $0.810\!\rightarrow\!0.972$, Stop $37.5\!\rightarrow\!81.2$, MAE $0.15\!\rightarrow\!0.11$). Overall, these results suggest that a pretrained progress predictor, when finetuned on a small amount of real data, transfers well to common scene shifts, supporting its use as a reliable signal for classifier guidance.

\subsection{Visualization}

Fig.~\ref{fig:visualization} visualizes three representative examples drawn from both simulation and realistic scenarios. Starting from the same time point, we present the achieved robotic states (particularly the progress scores) after performing same number of action tokens, with or without the proposed classifier guidance. The curves of progress scores across the entire operations are also displayed in the rightmost column in Fig.~\ref{fig:visualization}, which further validate the effectiveness of \vla.

\section{Conclusion}
\label{sec:conclusion}

We presented \vla, a progress-guided diffusion policy for robotic manipulation that adds an explicit progress signal to diffusion-based action generation. Its progress estimator, built on pretrained visual features, remains robust under \emph{original}, \emph{lighting-shift}, and \emph{novel-object} settings. Classifier guidance steers diffusion sampling with progress gradients, improving both success rates and execution efficiency, while RL finetuning further improves the robustness of both the evaluator and the policy.

\newpage
{
\bibliographystyle{plainnat}
\bibliography{references}
}

\clearpage
\appendices

\section{Implementation Details}

\subsection{Progress Estimator}

\label{sec:pp_arch}

The progress estimator is a compact cross-attention regressor that maps the instruction
\(\ell\), the start observation \(o_0\), and the current observation \(o_t\) to a scalar progress score
\(\hat p\in[0,1]\).
We first extract frozen pretrained features: visual patch tokens from the pretrained DINOv2\cite{oquab2023dinov2} and text tokens from the pretrained CLIP model (OpenAI CLIP ViT-L/14) \cite{radford2021learning}.
All visual and text tokens are then projected into a shared embedding space.
A lightweight cross-attention stack with residual connections and LayerNorm (i) aligns language with the current observation and
(ii) encodes start-to-current changes.
Finally, token features are mean-pooled, fused by a small MLP, and passed through a sigmoid head to predict \(\hat p\).

Concretely, let \(S\), \(C\), and \(T\) denote the projected (and role-embedded) start-frame visual tokens,
current-frame visual tokens, and instruction tokens, respectively.
Three residual cross-attention updates are applied:
(1) attend from \(T\) to \(C\) to inject current visual context into the language stream;
(2) attend from \(C\) to \(S\) to capture start-to-current changes; and
(3) attend from the current-conditioned visual tokens to \(T\) to obtain a language-conditioned visual representation.
The resulting streams are mean-pooled, concatenated, and passed through a fusion MLP and sigmoid head to obtain \(\hat p\).

The estimator uses 768-dim DINO patch features and 768-dim CLIP text features, each projected to width 1024,
with 8 attention heads, dropout 0.1, and a 6-block Spatio-Temporal Transformer backbone~\cite{xu2020spatial}.
The prediction head is an MLP with hidden size 512 and sigmoid output.
The model is pretrained with a total budget of at most \(400\) H20-hours, achieving competitive offline progress estimation and providing a reliable signal for downstream guidance.

\subsection{World Model}
The world model follows the UniVLA world-model architecture~\cite{bu2025univla}. Given observations \(o_t\) and \(o_{t+N}\), the pretrained DINOv2\cite{oquab2023dinov2} extract frozen  features \(F_t\) and \(F_{t+N}\), and model
latent action \(a^z\) in feature space (instead of pixel space) for robustness to appearance changes.

The encoder (inverse dynamics) maps \((F_t, F_{t+N})\) to a latent action \(a^z\), and the decoder (forward dynamics)
predicts future features conditioned on \(F_t\) and \(a^z\):
\begin{equation}
a^z = E(F_t, F_{t+N}), \qquad \hat F_{t+N} = D(F_t, a^z).
\label{eq:world_model}
\end{equation}
A VQ bottleneck is used for latent-action discretization before passing \(a^z\) to downstream policy modules.

In our experiments, both the encoder and decoder adopt a Spatio-Temporal Transformer backbone~\cite{xu2020spatial}, with hidden width \(768\), \(12\) Transformer layers, and \(12\) attention heads.
The latent action is parameterized with dimension \(128\).
Training minimizes an \(L_2\) feature reconstruction loss with standard VQ commitment/codebook terms, plus the KL
regularizer in Eq.~(5) of the main paper to improve latent stability and reduce distribution drift.

\subsection{Noise-conditioned evaluator}
\label{sec:q_sa_suppl}
Following the main-paper evaluator definition, candidate latent-action chunks are scored by applying the progress
estimator to world-model-imagined futures. For state \(s=(\ell,o_0,o_t)\) and latent-action chunk \(a\), the
task-aware score is
\(
Q(s,a)=P\!\big(\ell,o_0,D(o_t,a)\big).
\)

To apply this evaluator inside diffusion sampling, we use a noise-conditioned form
\begin{equation}
Q_\tau(x_\tau,\tau,s)
= P\!\big(\ell,o_0,D(o_t,x_\tau,\tau)\big),\ \tau\in\{0,\ldots,1000\},
\label{eq:q_tau_def}
\end{equation}
which is differentiable w.r.t.\ \(x_\tau\) and provides classifier-guidance gradients.

The base world model \(D(o_t,a)\) is \(\tau\)-agnostic. For evaluator prediction during classifier
guidance only, a lightweight \(\tau\)-conditioning branch is introduced and a guidance-time variant
\(D(o_t,x_\tau,\tau)\) is used. Specifically, \(\tau\) is encoded by a sinusoidal embedding
\(\mathbf{e}_\tau=\mathrm{TimeEmb}(\tau)\) and concatenated with the noisy latent action token \(x_\tau\) and current DINO features \(\mathbf{P}_t\):
\begin{equation}
\mathbf{X}_\tau=[a^z_\tau;\mathbf{e}_\tau;\mathbf{P}_t].
\label{eq:evaluator_tokens}
\end{equation}
A Transformer decoder predicts future DINO features from \(\mathbf{X}_\tau\), which are used in Eq.~\eqref{eq:q_tau_def}.
This \(\tau\)-conditioning stabilizes guidance gradients across noise levels and does not change the base world-model formulation.

To improve evaluator consistency from high-noise to low-noise steps (\(\tau:1000\!\rightarrow\!0\)), the world model and
progress estimator are jointly finetuned on expert demonstrations under the same noise-conditioning as in Eq.~\eqref{eq:q_tau_def},
then distilled into a noise-aware evaluator with a total compute budget of \(160\) H20-hours for classifier guidance.

\subsection{Latent Action Expert and Action Decoder}
\label{sec:latent_expert_action_decoder}

The policy is implemented as a two-stage diffusion system that factorizes planning into
(i) denoising a latent-action chunk and (ii) decoding the latent chunk into an executable action chunk.
Both stages adopt the LLaMA2-style diffusion-transformer architecture of \cite{hou2025dita}:
the observation is tokenized into a compact visual token sequence, concatenated with language conditioning,
and processed by a lightweight Transformer backbone. Each network is trained in the standard
variance-preserving diffusion setting to predict the noise residual (\emph{epsilon prediction}).

\subsubsection{Latent Action Expert}
The latent-action expert is a diffusion model operating in a compact latent-action space.
It takes the current observation and instruction as conditioning and iteratively denoises a noisy latent variable
to produce a latent-action chunk that captures task-relevant high-level intent.
A fine-grained diffusion schedule with $1000$ training timesteps is used for latent denoising,
which provides smoother intermediate noise levels for guidance and refinement.

\subsubsection{Action Decoder}
The action decoder is another diffusion model that generates the executable action chunk.
It is conditioned on the observation and instruction, and additionally takes the (noisy) latent-action variable
as an explicit conditioning signal throughout denoising.
Intuitively, the latent-action chunk represents an action plan in the visual (image) space, and the decoder translates it into
low-level actions consistent with the robot embodiment.
A shorter diffusion schedule with $100$ training timesteps is used for action denoising,
which reduces inference cost while retaining sufficient fidelity.

\subsection{Two-stage inference with latent warm-start}
\label{sec:two_stage_infer}

Inference uses two coupled diffusion processes: a \emph{latent-action} diffusion (fine schedule) and an \emph{action-chunk} diffusion (coarse schedule).
The latent-action diffusion uses \(\Tau_z=1000\) training timesteps, while the action-chunk diffusion uses \(\Tau_a=100\).
At test time, the noise scales are aligned via a two-stage procedure.
This two-stage design is motivated by differing denoising difficulty: decoding a coherent latent plan benefits from a longer schedule, while the action chunk is simpler and can be generated faithfully with \(\Tau_a{=}100\) steps.

\subsubsection{Stage 1: latent warm-start to the \texorpdfstring{$\Tau_a{=}100$}{Tau\_a=100} noise scale}
The latent-action denoiser is first run alone under the \(\Tau_z{=}1000\) schedule, but only for the tail segment that corresponds to timesteps \(\tau \ge \Tau_a\).
Starting from Gaussian noise \(x^{z}_{\tau=\Tau_z}\sim\mathcal{N}(0,I)\), the following iterative updates are applied:
\begin{equation}
\begin{aligned}
\epsilon^{z}_{\tau} &= \epsilon_{\theta}^{z}(x^{z}_{\tau},\, \tau,\, s), \\
\tilde\epsilon^{z}_{\tau} &=
\begin{cases}
\epsilon^{z}_{\tau}-\sigma_{\tau}^{z}\nabla_{x_{\tau}^z}Q_\tau(x_{\tau}^z,\tau,s), & \text{if CG on},\\
\epsilon^{z}_{\tau}, & \text{otherwise},
\end{cases}\\
x^{z}_{\tau-1} &\leftarrow \mathrm{Step}_{z}\!\left(x^{z}_{\tau},\, \tilde\epsilon^{z}_{\tau},\, \tau\right),
\qquad \forall\, t \in \{\Tau_z, \Tau_z\!-\!1,\ldots, \Tau_a\},
\end{aligned}
\label{eq:latent_warmstart}
\end{equation}
yielding a partially denoised latent \(x^{z}_{\Tau_a}\) at the noise scale compatible with the action diffusion.

\subsubsection{Stage 2: joint denoising of latent and action chunks}
A \(\Tau_a{=}100\)-step denoising loop is then run to jointly update the latent variable and the action chunk.
An action-chunk noise \(x^{a}_{\Tau_a}\sim\mathcal{N}(0,I)\) is initialized and the following coupled updates are applied for \(\tau=\Tau_a,\ldots,1\):
\begin{align}
\epsilon^{z}_{\tau} &= \epsilon_{\theta}^{z}(x^{z}_{\tau},\, \tau,\, s), \nonumber\\
\tilde\epsilon^{z}_{\tau} &=
\begin{cases}
\epsilon^{z}_{\tau}-\sigma_{\tau}^{z}\nabla_{x_{\tau}^z}Q_\tau(x_{\tau}^z,\tau,s), & \text{if CG on},\\
\epsilon^{z}_{\tau}, & \text{otherwise},
\end{cases}\nonumber\\
x^{z}_{\tau-1} &\leftarrow \mathrm{Step}_{z}\!\left(x^{z}_{\tau},\, \tilde\epsilon^{z}_{\tau},\, \tau\right), \nonumber\\
\epsilon^{a}_{\tau} &= \epsilon_{\theta}^{a}(x^{a}_{\tau},\, x^{z}_{\tau},\, \tau,\, s), \nonumber\\
x^{a}_{\tau-1} &\leftarrow \mathrm{Step}_{a}\!\left(x^{a}_{\tau},\, \epsilon^{a}_{\tau},\, \tau\right).
\label{eq:joint_denoise}
\end{align}
Here, \(\epsilon_{\theta}^{z}\) is the latent-action denoiser and \(\epsilon_{\theta}^{a}\) is the action decoder conditioned on the updated latent \(x^{z}_{\tau-1}\); \(\mathrm{Step}_{z}\) and \(\mathrm{Step}_{a}\) denote one DDIM scheduler step. CG is shorthand for classifier guidance.

After Stage 2, \(x^{a}_{0}\) is taken as the predicted action chunk to execute.
This two-stage procedure ensures that the latent plan is first brought to a compatible noise scale and then refined jointly with low-level action denoising, so that latent improvements immediately influence action updates within the same diffusion trajectory.

\section{Real-World Experiments}
\label{sec:supp-realworld}

\begin{figure}[h]
    \centering
    \setlength{\fboxsep}{6pt}
        \includegraphics[width=0.95\linewidth]{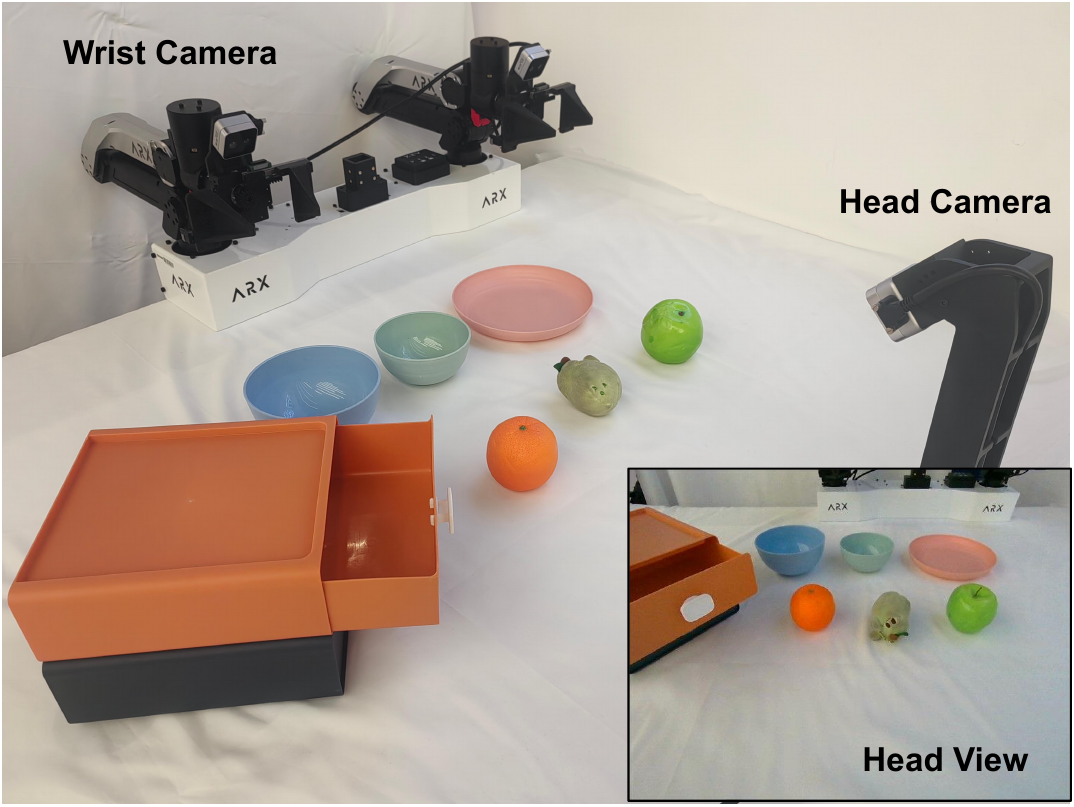}%
    \caption{\textbf{Real-world scenes.} All real-robot trials are executed using only the right arm for manipulation, while the left arm remains stationary throughout the rollout.}
    \label{fig:real_world_scene}
\end{figure}

\subsection{Robot platform and sensors}
\label{sec:robot_platform_sensors}
Real-world experiments are conducted on an ARX AC-One dual-arm platform equipped with two X5 arms and ARX G2
parallel grippers.
The sensory setup consists of two Intel RealSense D405 RGB-D cameras: one wrist-mounted on the end-effector
and one fixed as a stationary third-person ``head-view'' camera (see Fig.~\ref{fig:real_world_scene}).
Although both cameras support depth measurements, we do not use depth; the policy takes RGB images as input.

\subsection{Policy execution and evaluation}
\begin{figure}[h]
    \centering
    \includegraphics[width=0.95\linewidth]{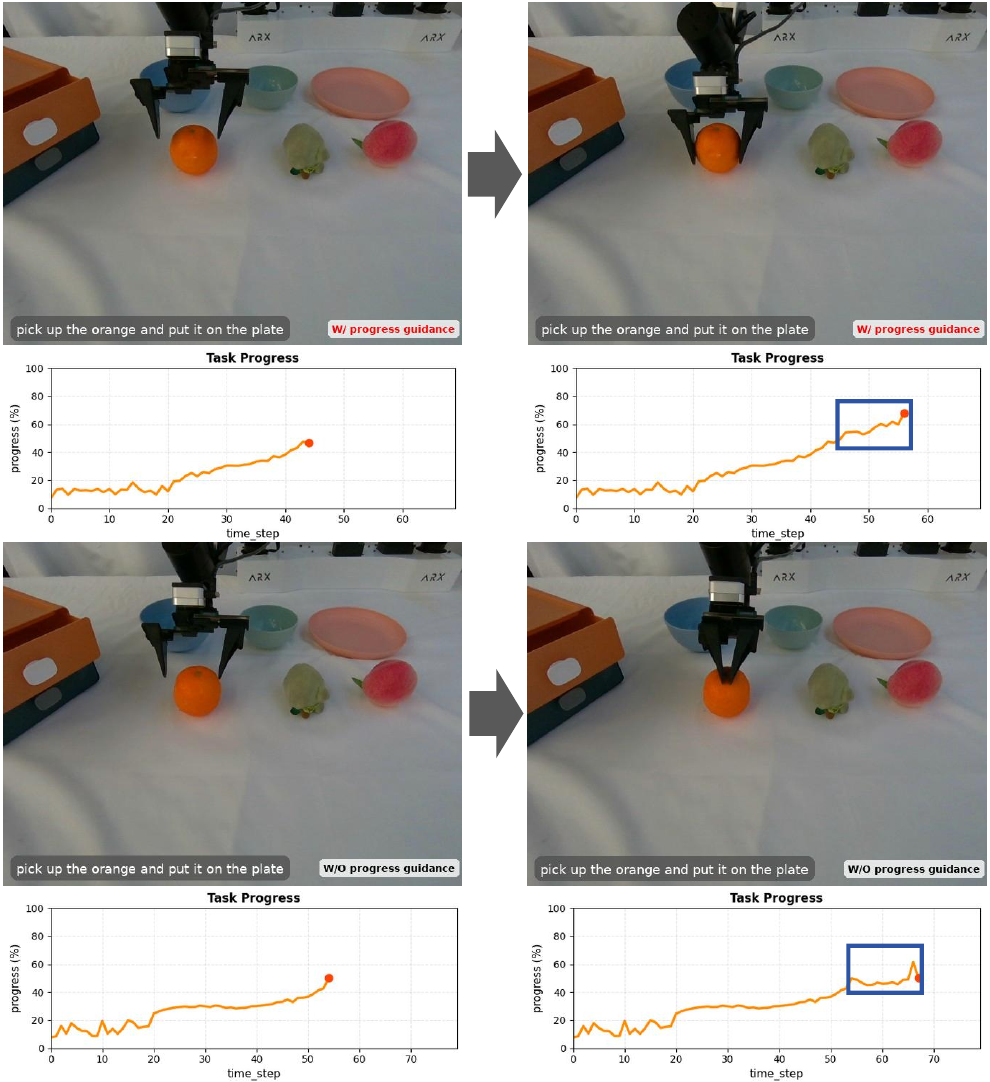}%
    \caption{\textbf{Progress estimator traces with/without progress guidance.} The rollout corresponds to the instruction ``pick up the orange and put it on the plate''. Top: with progress guidance. Bottom: without progress guidance. The orange curve shows the predicted task progress over time.}
    \label{fig:progress_trace_cg}
\end{figure}

Fig.~\ref{fig:progress_trace_cg} visualizes the progress estimator outputs during a real-robot rollout for the same instruction under two inference settings: \emph{without} progress guidance and \emph{with} progress guidance.
The predicted progress (in \%) is plotted against the rollout timestep, where higher values indicate that the evaluator believes the policy is closer to completing the task.

In Fig.~\ref{fig:progress_trace_cg}, the \textbf{top two panels} show rollouts \emph{with} progress guidance, while the \textbf{bottom two panels} show rollouts \emph{without} progress guidance.
With progress guidance, the gripper successfully closes on and lifts the orange, and the highlighted progress segment (blue box) increases \emph{monotonically}, indicating consistent evaluator-measured advancement.
Without progress guidance, the gripper fails to properly grasp the orange, and the highlighted progress segment exhibits noticeable \emph{oscillations}, suggesting unstable advancement signals and dithering during execution.

This is consistent with classifier guidance: during diffusion sampling, the evaluator-score gradient biases updates toward latent-action chunks predicted to achieve higher progress.
Consequently, the guided policy selects task-advancing actions more reliably, reducing both steps and unnecessary motion.

\subsection{Progress Estimator generalization}
\label{sec:pp_pretrain_vs_scratch}
A \emph{pretrained+finetuned} progress estimator is compared with a \emph{from-scratch} one under two appearance shifts: \emph{lighting change} and \emph{novel objects}.

\subsubsection{Lighting shift}
\begin{figure}[h]
    \centering
    \includegraphics[width=0.95\linewidth]{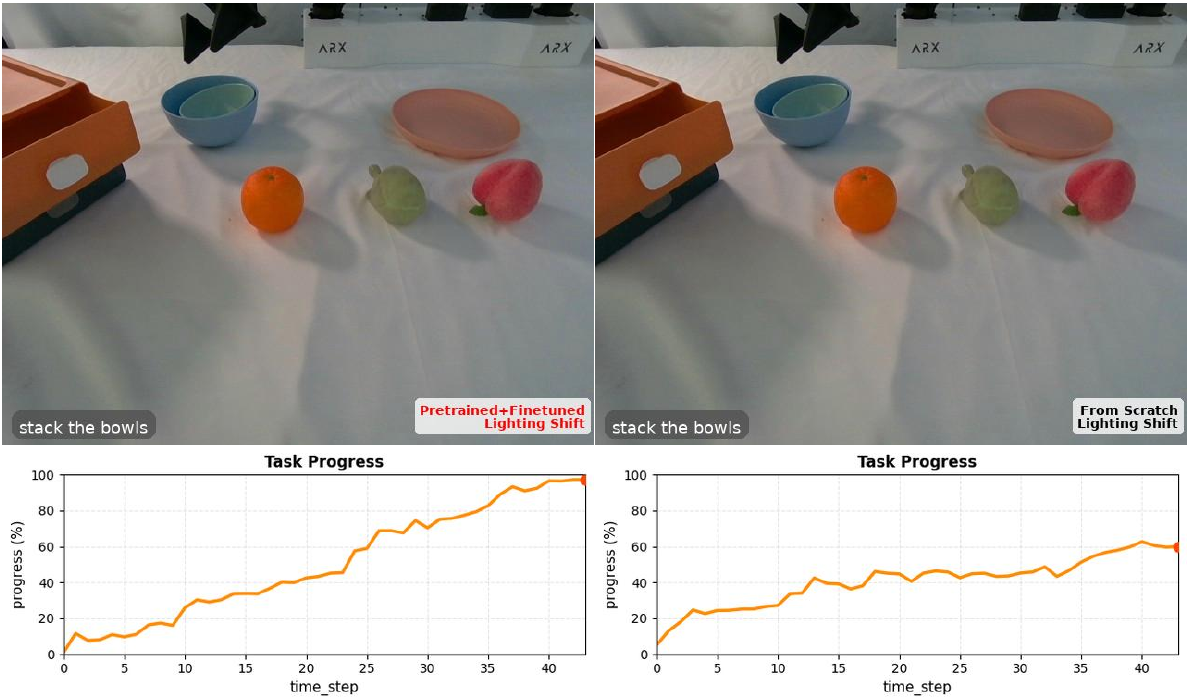}%
    \caption{\textbf{Progress traces under lighting shift.} The rollout is an expert-collected trajectory for evaluating the progress estimator (instruction: ``stack the bowls''). Left: Pretrained+Finetuned. Right: From Scratch.}
    \label{fig:pp_pretrain_vs_scratch}
\end{figure}

Fig.~\ref{fig:pp_pretrain_vs_scratch} compares progress traces under a lighting-shift setting for the same real-robot instruction.

Under lighting shift, the pretrained+finetuned estimator (left) stays smooth and near-monotonic, reaching high progress in fewer steps.
The from-scratch estimator (right) rises more slowly and plateaus in the later stage, making completion harder to identify.

\subsubsection{Novel-object shift}
\begin{figure}[h]
    \centering
    \includegraphics[width=0.95\linewidth]{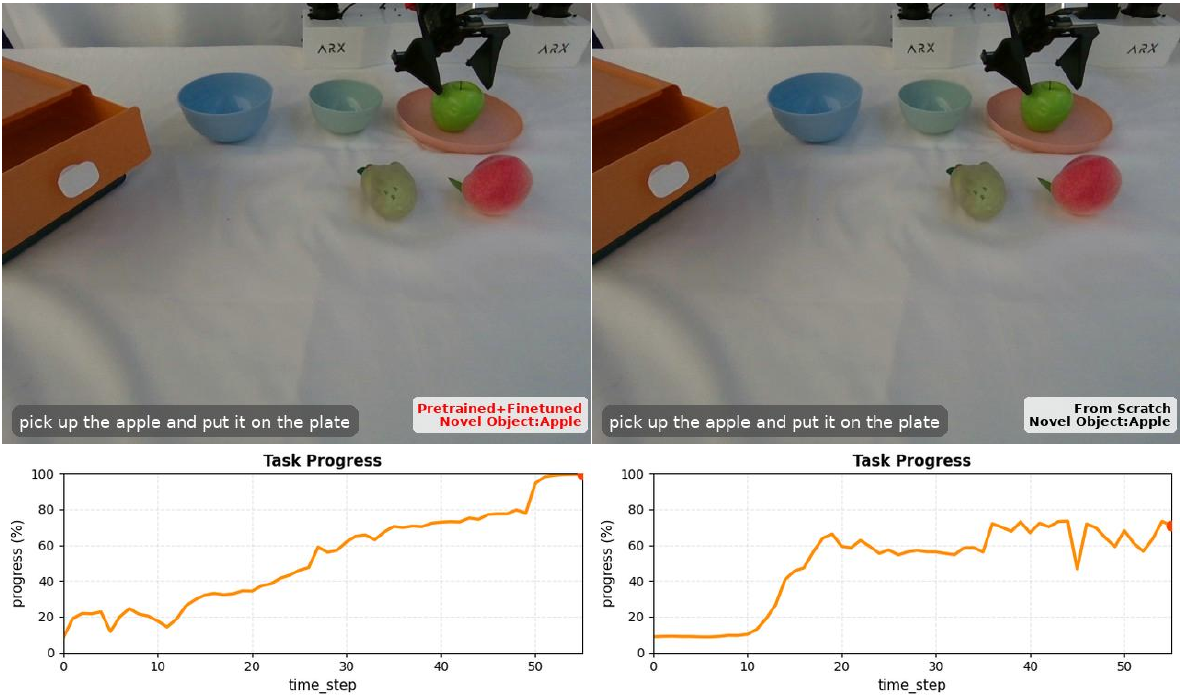}%
    \caption{\textbf{Progress traces under novel objects.} The rollout is an expert-collected trajectory for evaluating the progress estimator (instruction: ``pick up the apple and put it on the plate''). Left: Pretrained+Finetuned. Right: From Scratch.}
    \label{fig:pp_pretrain_vs_scratch_novel}
\end{figure}

A \emph{novel-object} setting is further tested, where the target object differs from training.
Fig.~\ref{fig:pp_pretrain_vs_scratch_novel} shows progress traces for an unseen object manipulation instruction.

With novel objects, the pretrained+finetuned estimator (left) remains smooth/near-monotonic and approaches high progress near the end.
The from-scratch estimator (right) fluctuates more and gives a less separable near-completion region, making it harder to judge whether the task is almost done.

Pretraining + finetuning improves robustness to appearance shifts, yielding more accurate progress prediction and a clearer terminal signal.

\subsection{Additional video results}
More qualitative visualizations are provided in the supplementary \texttt{videos/} folder.
Progress-guidance videos are in \texttt{videos/Guidance/} and are named by the corresponding instruction.
Progress-estimator videos are in \texttt{videos/Estimator/} and are named \texttt{novel\_object} and \texttt{light\_shifting}.

\section{Proof of Policy Improvement}
\label{sec:supp-proof}

\subsection{Task-aware score from the evaluator}
Following Eq.~(16) in the main paper, let the state be $s=(\ell,o_0,o_t)$ and the (latent) action be $a$.
Given a learned evaluator (world model + progress estimator), the task-aware score is defined as
\begin{equation}
Q(s,a)\ =\ P\!\big(\ell,o_0,\ D(o_t,a)\big),
\label{eq:paper_eq16_recap}
\end{equation}
i.e., the predicted progress after applying $a$ through the world model $D$.

\subsection{KL-constrained improvement}
As in Eq.~(17) in the main paper, progress maximization is cast as a KL-regularized policy improvement:
\begin{equation}
\begin{aligned}
\pi^\star(\cdot|s)
&=\arg\max_{\pi(\cdot|s)}\ \mathbb{E}_{a\sim\pi(\cdot|s)}\!\big[Q(s,a)\big], \\
\text{s.t.} &\quad \mathrm{KL}\!\big(\pi(\cdot|s)\,\|\,\pi_0(\cdot|s)\big)\le \varepsilon .
\end{aligned}
\label{eq:paper_eq17_recap}
\end{equation}
To solve the KL-constrained problem (Boltzmann form), let us first
introduce a Lagrange multiplier $\alpha>0$ and write the Lagrangian
\begin{equation}
\mathcal{L}(\pi;\alpha)
=\mathbb{E}_{a\sim\pi(\cdot|s)}\!\big[Q(s,a)\big]
-\alpha\Big(\mathrm{KL}\!\big(\pi(\cdot|s)\,\|\,\pi_0(\cdot|s)\big)-\varepsilon\Big).
\label{eq:kl_lagrangian}
\end{equation}
Taking the functional derivative w.r.t.\ $\pi(a|s)$ and enforcing normalization yields the unique optimum
\begin{equation}
\begin{aligned}
\pi^\star(a|s)
&=\frac{1}{Z(s)}\,\pi_0(a|s)\exp\!\left(\frac{1}{\alpha}Q(s,a)\right), \\
Z(s)
&=\int \pi_0(a|s)\exp\!\left(\frac{1}{\alpha}Q(s,a)\right)\,da ,
\end{aligned}
\label{eq:boltzmann_policy_detail}
\end{equation}
which matches Eq.~(18) in the main paper. It implies that the log-density differs by an additive energy term:
\begin{equation}
\begin{aligned}
\log \pi^\star(a|s)
&=\log \pi_0(a|s)+\frac{1}{\alpha}Q(s,a)-\log Z(s), \\
\nabla_a \log \pi^\star(a|s)
&=\nabla_a \log \pi_0(a|s)+\frac{1}{\alpha}\nabla_a Q(s,a),
\end{aligned}
\label{eq:guided_score_on_a}
\end{equation}
since $Z(s)$ does not depend on $a$.

\subsection{Instantiating $\pi_0$ as a VP (Variance-Preserving) diffusion policy over latent actions}
The action variable $a$ is now set to the diffusion latent $x_0$, and the denoising variable
$x_\tau$ at diffusion step $\tau$. Under the VP forward process,
\begin{equation}
x_\tau=\sqrt{\bar\alpha_\tau}\,x_0+\sigma_\tau \epsilon, 
\quad \epsilon\sim\mathcal{N}(0,I),\quad \sigma_\tau^2=1-\bar\alpha_\tau .
\label{eq:vp_forward_recap}
\end{equation}
The diffusion policy is parameterized by an $\epsilon$-predictor $\epsilon_\theta(x_\tau,\tau,s)$.
For VP diffusion, the score of the base model satisfies
\begin{equation}
s_\theta(x_\tau,\tau,s)\ \triangleq\ \nabla_{x_\tau}\log p_\theta(x_\tau|s)
\ \approx\ -\frac{1}{\sigma_\tau}\,\epsilon_\theta(x_\tau,\tau,s),
\label{eq:score_eps_vp_recap}
\end{equation}
where the approximation becomes exact when $\epsilon_\theta$ matches the conditional mean
$\mathbb{E}[\epsilon|x_\tau,\tau,s]$.

\subsection{Classifier guidance on the denoising variable $x_\tau$}
To apply the KL-derived improvement during sampling, the evaluator is used to define a noise-aware score
$Q_\tau(s,x_\tau)$ (the evaluator takes $(x_\tau,\tau,s)$ as input).
Applying Eq.~\eqref{eq:guided_score_on_a} to the denoising variable gives the guided score
\begin{equation}
\begin{aligned}
s^\star(x_\tau,\tau,s)
&\triangleq \nabla_{x_\tau}\log p^\star(x_\tau|s) \\
&= s_\theta(x_\tau,\tau,s)+\frac{1}{\alpha}\nabla_{x_\tau}Q_\tau(s,x_\tau).
\end{aligned}
\label{eq:guided_score_xtau}
\end{equation}

\subsection{Converting guided score to a guided $\epsilon$ target (Eq.~(19))}
Define a guided noise target $\tilde\epsilon$ by $s^\star(x_\tau,\tau,s)=-\sigma_\tau^{-1}\tilde\epsilon$.
Combining Eq.~\eqref{eq:score_eps_vp_recap} and Eq.~\eqref{eq:guided_score_xtau} yields
\begin{equation}
\begin{aligned}
-\frac{1}{\sigma_\tau}\tilde\epsilon
&=-\frac{1}{\sigma_\tau}\epsilon_\theta(x_\tau,\tau,s)+\frac{1}{\alpha}\nabla_{x_\tau}Q_\tau(s,x_\tau), \\
\tilde\epsilon
&=\epsilon_\theta(x_\tau,\tau,s)-\frac{\sigma_\tau}{\alpha}\nabla_{x_\tau}Q_\tau(s,x_\tau),
\end{aligned}
\label{eq:guided_eps_eq19}
\end{equation}
which is exactly Eq.~(19) in the main paper (up to notation).

Finally, the guided direction is distilled into the denoiser by minimizing the standard denoising objective
with the guided target:
\begin{equation}
\mathcal{L}_{\text{policy}}
=\mathbb{E}\!\left[\left\|\tilde\epsilon-\epsilon_\theta(x_\tau,\tau,s)\right\|_2^2\right],
\label{eq:policy_loss_eq20}
\end{equation}
matching Eq.~(20) in the main paper.

\end{document}